\documentclass{article}

\usepackage{arxiv}
\PassOptionsToPackage{numbers}{natbib}
\usepackage[utf8]{inputenc} 
\usepackage[T1]{fontenc}    
\usepackage{hyperref}       
\usepackage{url}            
\usepackage{booktabs}       
\usepackage{amsmath,amssymb,amsfonts}
\usepackage{nicefrac}       
\usepackage{microtype}      
\usepackage{lipsum}		
\usepackage{graphicx}
\usepackage{natbib}
\usepackage{doi}
\usepackage{tabularx}
\usepackage{textcomp}
\usepackage{hyperref} 
\usepackage{algorithmic}

\title{Vision-QRWKV: Exploring Quantum-Enhanced RWKV Models for Image Classification}

\date{} 					

\author{%
  \href{https://orcid.org/0000-0003-0807-0217}{\includegraphics[scale=0.06]{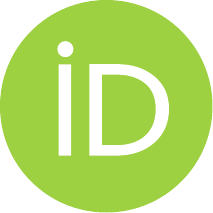}\hspace{1mm}Chi-Sheng Chen\thanks{Corresponding author}} \\
  Neuro Industry Research\\
  Neuro Industry, Inc.\\
  Boston, MA, USA \\
  \texttt{m50816m50816@gmail.com} \\
  \texttt{cchen34@bidmc.harvard.edu} \\
}

\renewcommand{\shorttitle}{\textit{arXiv} Template}

\hypersetup{
pdftitle={A template for the arxiv style},
pdfsubject={cs.CV},
pdfauthor={Chi-Sheng ~Chen},
pdfkeywords={Deep learning},
}

\begin{document}
\maketitle

\begin{abstract}
Recent advancements in quantum machine learning have shown promise in enhancing classical neural network architectures, particularly in domains involving complex, high-dimensional data. Building upon prior work in temporal sequence modeling, this paper introduces Vision-QRWKV, a hybrid quantum-classical extension of the Receptance Weighted Key Value (RWKV) architecture, applied for the first time to image classification tasks. By integrating a variational quantum circuit (VQC) into the channel mixing component of RWKV, our model aims to improve nonlinear feature transformation and enhance the expressive capacity of visual representations.

We evaluate both classical and quantum RWKV models on a diverse collection of 14 medical and standard image classification benchmarks, including MedMNIST datasets, MNIST, and FashionMNIST. Our results demonstrate that the quantum-enhanced model outperforms its classical counterpart on a majority of datasets, particularly those with subtle or noisy class distinctions (e.g., ChestMNIST, RetinaMNIST, BloodMNIST). This study represents the first systematic application of quantum-enhanced RWKV in the visual domain, offering insights into the architectural trade-offs and future potential of quantum models for lightweight and efficient vision tasks. 
The code can be accessed on GitHub: \url{https://github.com/ChiShengChen/Vision_QRWKV}.
\end{abstract}

\keywords{Deep learning \and Image Classification \and RWKV \and Quantum Machine Learning }

\section{Introduction}

Deep neural networks have achieved remarkable success in image classification across various domains, including medical imaging~\cite{lai2023intraoperative}, fine-grained images~\cite{chen2025improving, chen2024food}, and natural scenes~\cite{chen2024necomimi}. While convolutional neural networks (CNNs) and vision transformers (ViTs)~\cite{vaswani2017attention, dosovitskiy2020image} remain dominant, alternative architectures such as RWKV~\cite{peng2023rwkv}, a hybrid of recurrent and feedforward mechanisms, have emerged as scalable, attention-free models suitable for both sequence and visual data. However, like many classical architectures, RWKV's expressiveness in complex feature spaces is ultimately constrained by the linearity and structure of its feedforward layers.

Recent research in quantum machine learning (QML)~\cite{biamonte2017quantum} has explored the integration of variational quantum circuits (VQCs) into deep models to enhance nonlinear representational power. In particular, our prior work introduced QuantumRWKV~\cite{chen2025qrwkv}, a hybrid quantum-classical extension of RWKV for time series forecasting, demonstrating improvements on chaotic and nonlinear temporal patterns. Building on that success, we now extend this architecture into the visual domain.

In this paper, we present Vision-QRWKV, the first application of a quantum-enhanced RWKV model to image classification tasks. Our architecture maintains the attention-free, time-mixing backbone of RWKV while embedding a trainable VQC into its channel-mixing (feedforward) layer. This design allows quantum computation to supplement classical processing by capturing richer nonlinear interactions in image representations.

To assess the efficacy of this hybrid approach, we conduct comprehensive experiments across 14 image classification datasets, including MedMNIST medical datasets~\cite{yang2023medmnist}, MNIST~\cite{deng2012mnist}, and FashionMNIST~\cite{xiao2017fashion}. For each dataset, we train and evaluate both the classical and quantum variants of RWKV under identical conditions and compare test accuracy, confusion matrices, and per-dataset improvements.

Our key findings are as follows:
\begin{itemize}
    \item The quantum RWKV model achieves consistent performance gains on datasets where class boundaries are blurred or subtle, such as ChestMNIST, RetinaMNIST, and BloodMNIST.
    \item In simpler or more structured datasets like MNIST and OrganAMNIST, performance is comparable, demonstrating that the quantum model does not compromise efficiency or stability.
    \item This study offers a novel benchmark for hybrid quantum-classical models in the image domain, setting the stage for future research in multimodal QML, medical imaging, and vision-language models.
\end{itemize}

\section{Related Work}

\subsection{RWKV and Attention-Free Models}
The RWKV architecture~\cite{peng2023rwkv} was recently introduced as an attention-free alternative to Transformers, combining recurrent-style time-mixing with feedforward channel mixing. Originally designed for long-context sequence modeling, RWKV achieves linear time and memory complexity while maintaining competitive performance across language and time-series domains. While prior applications of RWKV have focused on text and temporal data, its potential for vision tasks remains largely unexplored. Our work is the first to adapt RWKV---and its quantum-enhanced variant---to image classification.

\subsection{Quantum Neural Networks for Vision}
Quantum neural networks (QNNs) have been investigated as expressive models capable of learning complex data structures in high-dimensional Hilbert spaces~\cite{goto2021universal, chen2025quantum}. Notable applications include quantum classifiers, quantum convolutional neural networks (QCNNs), and quantum generative models. Recent work~\cite{chen2024qeegnet} has explored integrating quantum layers with classical encoders for brain imaging and electroencephalography (EEG) tasks and MNIST image generation~\cite{chen2025quantumgen}. However, very few studies have applied QNNs to standard or medical image classification tasks at scale. Our work extends this direction by integrating variational quantum circuits (VQCs) directly into a scalable image classification backbone.

\subsection{Hybrid Quantum-Classical Architectures}
Hybrid models that combine classical neural layers with quantum components, typically via variational quantum circuits, have gained popularity due to their compatibility with current Noisy Intermediate-Scale Quantum (NISQ) hardware~\cite{de2022survey}. Prior work has shown that even shallow VQCs can enhance non-linear learning capacity in low-data or chaotic regimes~\cite{chen2025qrwkv}. Our architecture adopts a similar philosophy, replacing the feedforward network in RWKV with a quantum-enhanced counterpart to explore potential gains in visual domains.

\begin{figure}[ht]
\centering
\includegraphics[width=\linewidth]{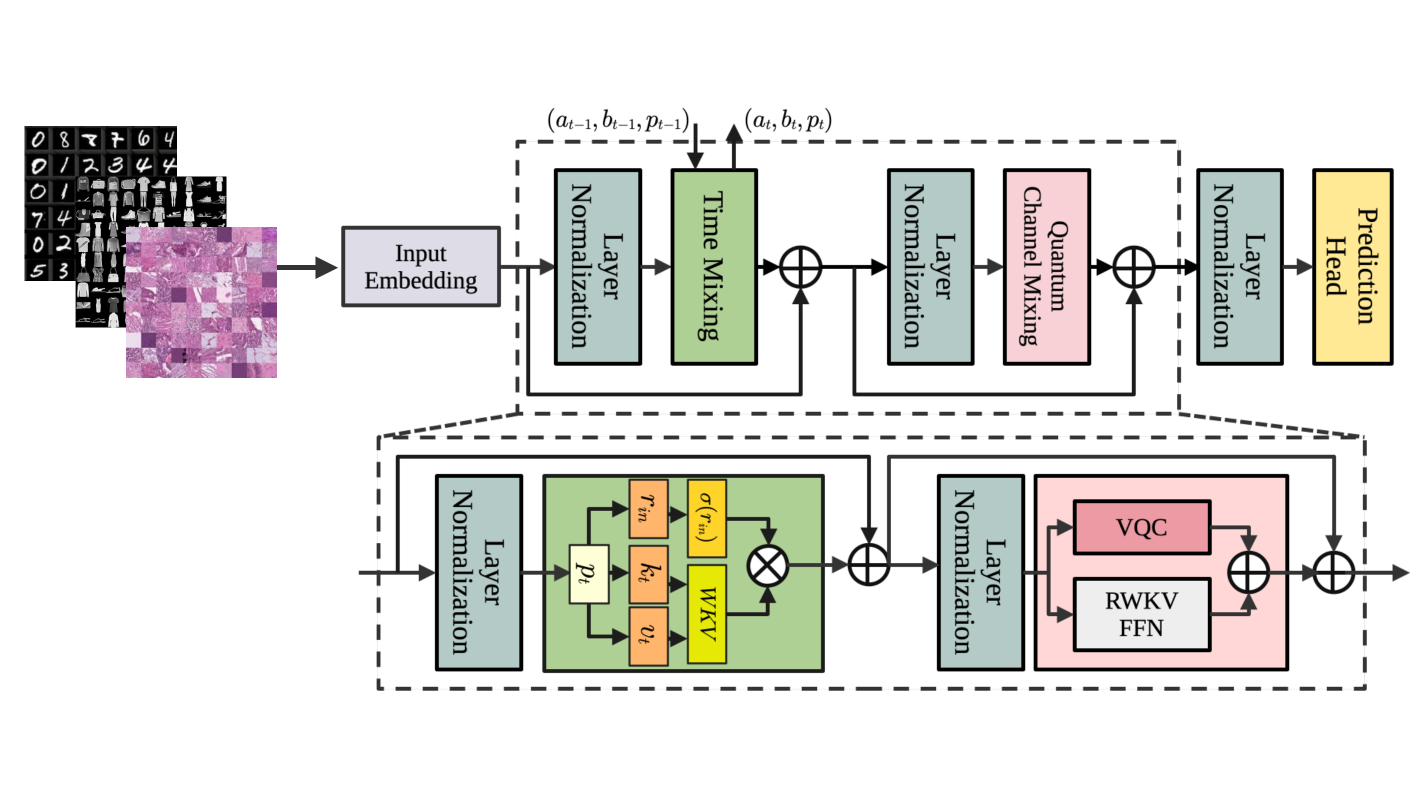}
\caption{%
Overview of the Vision QuRWKV architecture. Input images (e.g., MNIST, FashionMNIST, and MedMNIST variants) are first embedded into token representations. Each block in the model consists of a \textit{Time Mixing} module (green), which performs recurrent-style memory updates using weighted key-value operations, and a \textit{Quantum Channel Mixing} module (pink), which combines classical feedforward transformation with a variational quantum circuit (VQC). Layer normalization is applied before each submodule. The final output is passed through a prediction head for classification. The bottom panel details the internal operations of each submodule. All quantum layers are fully differentiable and trained end-to-end.}
\label{fig:vision_qrwkv_architecture}
\end{figure}

\section{Methodology}

\subsection{Overview of RWKV Architecture}
The original RWKV model consists of two main components per layer: a \textit{time-mixing} module, which processes sequential dependencies using exponential decay weights, and a \textit{channel-mixing} module, which resembles a position-wise feedforward network. In the classical setting, the channel mixing operation is defined as:
\begin{equation}
    h = \sigma(r) \odot W_2 \left(\text{ReLU}(W_1 x)\right)
\end{equation}
where $r$ is a learned receptance vector, and $W_1$, $W_2$ are trainable projection matrices.

\subsection{Quantum-Enhanced Channel Mixing}
We extend the channel mixing module by incorporating a variational quantum circuit (VQC). The classical hidden state is first projected into a lower-dimensional quantum embedding space:
\begin{equation}
    x_q = W_q x
\end{equation}
Each element of $x_q$ is encoded via angle embedding into rotation gates (e.g., RX or RY) applied to a quantum circuit with $n$ qubits. The VQC consists of $L$ entangling layers using CNOT gates arranged in a ladder pattern. The circuit outputs the expectation values of the Pauli-Z operators:
\begin{equation}
    z_i = \langle \psi | Z_i | \psi \rangle
\end{equation}
These are then linearly projected and fused with the classical FFN output:
\begin{equation}
    \text{QuantumMix}(x) = \sigma(r) \odot \left( W_2(\text{ReLU}(W_1 x)) + W_o z \right)
\end{equation}
where $W_o$ maps the quantum output back to the model's embedding dimension.

\subsection{Vision-QRWKV Model}
To apply RWKV to image classification, we flatten each image into a 1D token sequence and apply multiple stacked layers of the hybrid RWKV blocks:
\begin{equation}
    x^{(l+1)} = x^{(l)} + \text{TimeMix}(\text{LN}(x^{(l)})) + \text{QuantumMix}(\text{LN}(x^{(l)} + \text{TimeMix}))
\end{equation}
The final output is aggregated via a global average pooling or classification head. All quantum layers are simulated using the PennyLane \texttt{default.qubit} backend, allowing gradient backpropagation through the full pipeline.

\section{Experiments}

\subsection{Datasets}
To evaluate the performance of the proposed Vision-QRWKV model, we conduct experiments on a diverse set of image classification benchmarks, including both standard and medical imaging datasets. Specifically, we include:

\begin{itemize}
    \item \textbf{Standard Datasets:} MNIST and FashionMNIST, containing 28$\times$28 grayscale images of handwritten digits and clothing items, respectively.
    \item \textbf{MedMNIST Collection}~\cite{yang2023medmnist}: A curated benchmark for lightweight biomedical image classification, comprising 12 datasets covering modalities such as histopathology, chest X-ray, dermoscopy, breast ultrasound, and retina scans. We select the following: BloodMNIST, TissueMNIST, OCTMNIST, PathMNIST, ChestMNIST, OrganAMNIST, OrganSMNIST, OrganCMNIST, DermaMNIST, PneumoniaMNIST, RetinaMNIST, and BreastMNIST.
\end{itemize}

All images are resized to 8$\times$8 resolution to maintain consistency across datasets and enable direct comparison with MNIST-scale models.

\subsection{Model Configuration}
Both classical and quantum RWKV models are implemented with the same architectural backbone. The only difference lies in the channel-mixing component:
\begin{itemize}
    \item \textbf{Classical RWKV:} Uses a two-layer feedforward network with ReLU activation.
    \item \textbf{Quantum RWKV:} Replaces the FFN with a hybrid branch combining a VQC and a classical MLP.
\end{itemize}

We use an embedding size of 768, 4 stacked RWKV blocks, and batch normalization. For the VQC, we simulate a 4-qubit circuit with a depth of 2 layers using the PennyLane \texttt{default.qubit} backend. Quantum parameters are trained end-to-end using backpropagation.

\subsection{Training Setup}
All models are trained using the Adam optimizer with a learning rate of $1 \times 10^{-3}$ and batch size of 64. Each dataset is trained for 30 epochs. No data augmentation or regularization is applied to ensure consistency across models. Each experiment is repeated with 3 different random seeds, and the mean test accuracy is reported.

\subsection{Training Details}

All models are trained for 30 epochs. The optimizer used is Adam with a fixed learning rate of $10^{-3}$, and the batch size is set to 64 for all experiments. No learning rate scheduling, dropout, or early stopping is applied, allowing consistent comparison between classical and quantum models under identical training regimes.

The QuantumRWKV model incorporates 4 qubits with a circuit depth of 2. Each input vector is encoded using angle embedding into rotation gates, followed by two entangling layers composed of CNOT gates in a ladder topology. The quantum circuit is simulated using the \texttt{default.qubit} backend provided by PennyLane, and gradients are propagated using fully differentiable quantum-classical backpropagation.

All experiments are conducted on a single NVIDIA A100 GPU with 40GB of memory. Although the quantum-enhanced models introduce additional simulation overhead, the total training configurations, number of updates, and evaluation metrics are aligned across both model variants for fairness.

\section{Results}

\subsection{Overall Performance Comparison}
We compare the classification accuracy of classical and quantum-enhanced RWKV models across 14 image datasets, including both general-purpose and biomedical benchmarks. Table~\ref{tab:accuracy} summarizes the test accuracy of both models along with an indicator of whether the quantum model outperformed the classical counterpart.

Overall, the QuantumRWKV model outperforms the classical variant on 8 out of 14 datasets. Notable improvements are observed on datasets with subtle or noisy visual patterns, such as \textit{BloodMNIST}, \textit{ChestMNIST}, and \textit{RetinaMNIST}. On simpler datasets such as \textit{MNIST}, or structure-specific datasets such as \textit{OrganAMNIST}, the performance of the quantum model is comparable or slightly lower.

\begin{table}[ht]
\centering
\caption{Test Accuracy (\%) Comparison Between Classical and Quantum RWKV}
\label{tab:accuracy}
\begin{tabular}{lrrl}
\toprule
       Dataset &  Classical Accuracy (\%) &  Quantum Accuracy (\%) & Quantum Better? \\
\midrule
         MNIST~\cite{deng2012mnist} &                   96.49 &                 96.12 &             No  \\
    BloodMNIST~\cite{yang2023medmnist} &                   91.32 &                 92.22 &             Yes \\
  FashionMNIST~\cite{xiao2017fashion} &                   85.56 &                 86.08 &             Yes \\
PneumoniaMNIST~\cite{yang2023medmnist} &                   83.81 &                 84.78 &             Yes \\
   BreastMNIST~\cite{yang2023medmnist} &                   77.56 &                 77.56 &           Equal \\
    ChestMNIST~\cite{yang2023medmnist} &                   74.44 &                 77.26 &             Yes \\
   OrganAMNIST~\cite{yang2023medmnist} &                   78.24 &                 76.27 &              No \\
   OrganCMNIST~\cite{yang2023medmnist} &                   75.60 &                 74.66 &              No \\
    DermaMNIST~\cite{yang2023medmnist} &                   71.97 &                 71.97 &           Equal \\
     PathMNIST~\cite{yang2023medmnist} &                   74.00 &                 71.11 &              No \\
   OrganSMNIST~\cite{yang2023medmnist} &                   60.45 &                 59.18 &              No \\
      OCTMNIST~\cite{yang2023medmnist} &                   55.80 &                 57.00 &             Yes \\
   TissueMNIST~\cite{yang2023medmnist} &                   55.43 &                 55.48 &             Yes \\
   RetinaMNIST~\cite{yang2023medmnist} &                   49.25 &                 53.75 &             Yes \\
\bottomrule
\end{tabular}
\end{table}

\section{Discussion}

The experimental results demonstrate that the integration of variational quantum circuits (VQCs) into the RWKV architecture can yield tangible benefits in image classification tasks, particularly within medical imaging domains. The QuantumRWKV model achieves higher test accuracy on over half of the evaluated datasets, most notably those characterized by subtle class boundaries and noise-sensitive features, such as \textit{RetinaMNIST}, \textit{BloodMNIST}, and \textit{ChestMNIST}.

These findings are consistent with prior observations in our earlier work on time-series forecasting, where QuantumRWKV was shown to outperform classical models in chaotic and nonlinear regimes. In the image domain, we hypothesize that the VQC's enhanced nonlinear representation power contributes to improved feature discrimination in visually ambiguous settings. The angle embedding and entanglement layers may help capture inter-feature correlations that classical linear layers fail to model effectively.

However, this quantum advantage is not universal. On simpler datasets like \textit{MNIST} and \textit{OrganAMNIST}, the performance difference between classical and quantum models is marginal or favors the classical variant. This suggests that the expressive capacity of VQCs may not always translate into better generalization, especially when the underlying decision boundaries are already well-modeled by classical architectures.

Furthermore, the additional computational overhead introduced by quantum circuit simulation is non-trivial. While feasible for small circuits and modest datasets, scalability remains a challenge in real-world deployment, especially without access to physical quantum hardware.

\section{Conclusion}

In this work, we introduced Vision-QRWKV, the first quantum-enhanced RWKV model applied to image classification tasks. By embedding a variational quantum circuit into the channel mixing pathway of RWKV, we created a hybrid architecture capable of leveraging quantum nonlinearity within a scalable, attention-free framework.

Our experiments across 14 datasets---including standard and medical imaging benchmarks---demonstrate that the quantum-enhanced model achieves superior performance on tasks involving subtle or noisy visual features. These results extend the utility of QuantumRWKV beyond time-series forecasting into the visual domain, providing early evidence that quantum-classical hybrid models can be effective for lightweight vision applications.


Our findings serve as a bridge between scalable classical vision architectures and the emerging expressiveness of quantum computing, opening new avenues for hybrid deep learning systems in healthcare and beyond.

\section*{Acknowledgment}
The authors would like to thank Neuro Industry, Inc. for generously providing cloud computing resources via Google Cloud Platform (GCP), which enabled the training and evaluation of the models presented in this work.

\bibliographystyle{unsrtnat}
\bibliography{references}  

\end{document}